# AutoFAQ - Frequently Asked Questions Generation

Nagasai Chandra
nachandr@calpoly.edu
California Polytechnic State University
San Luis Obispo, California

Teja Kalavakolanu
akalvako@calpoly.edu
California Polytechnic State University
San Luis Obispo, California

Michael Fekadu
mfekadu@calpoly.edu
California Polytechnic State University
San Luis Obispo, California

## ABSTRACT

FAQ (Frequently Asked Questions) documents are commonly used with text documents and websites to provide the important information in the form of question-answer pairs to either aid in reading comprehension or to provide a shortcut to the key ideas. We suppose that salient sentences from a given document serve as a good proxy for the answers to an aggregated set of FAQs from readers. We propose a system for generating FAQ documents that extract the salient questions and their corresponding answers from sizeable text documents scraped from the Stanford Encyclopedia of Philosophy [23]. We use existing text summarization, sentence ranking via the TextRank algorithm, and question generation tools to create an initial set of questions and answers. Finally, we apply some heuristics to filter out invalid questions. We use human evaluation [12] to rate the generated questions on grammar, whether the question is meaningful, and on the question's answerability within a summarized context. By human-evaluation on four articles, we find that grammar and answerability of questions generated have average ratings of 4.38 and 3.81, respectively. On average, participants thought 71% of questions were meaningful.

## KEYWORDS

Plato corpus, Transformers, Natural Language Understanding, Natural Language Generation, Neural Networks, Text Tagging, Question Generation, Page-Rank, Text-Rank, Summarization,

## 1 INTRODUCTION

Readers of an informative website may often skip through the critical concepts, hidden or spread out, into different web pages or fine print. In such cases, FAQs serve as a good start to guide their reading to focus on the right paragraph or to clarify confusing sections of text. In general, FAQ's are written manually by determining the frequencies of questions received from target readers about the document. In this work, we develop a system that can generate a set of salient questions and answers given a sizeable text document. We assume that a good set of predetermined FAQs (not depending on readers asking questions) contains the salient concepts of a given document. We consider the philosophical documents from Stanford Encyclopedia of Philosophy (SEP)[23] [1]. This corpus contains large text documents about philosophical concepts that have an average Flesch Reading ease score of about 50; thus, many readers may struggle to comprehend these articles quickly. We organized and analyzed our data using Kaggle. [2]

Our approach ?? to generate FAQs for such documents involves using *summarization* techniques to identify salient paragraphs, using *TextRank* algorithm which is a modification of *PageRank* to rank the information at sentence-level, and converting the highly ranked information into questions. Finally, we use four heuristics to finely filter the generated questions to guarantee that good question grammar and salience to the document. We relied on human judgment to evaluate the quality of generated FAQs based on different criteria like meaningfulness, the provided answer being informative (contains salient facts from a document), and quality of the grammar. Nema et al. (2018) [12] found that human judgement remains better than automated methods for evaluating question generation systems.

The four heuristics include a mandatory condition of using a question answering system to check whether the generated questions gives the answer that contains the blank part of the question. The other three heuristics are to check for decent grammar if the question has two $Wh$ words and checking if a named entity is present. If the questions fail any two heuristics among them, then it is filtered out.

To evaluate the generated questions, as proven by Nema et al. (2018)

In the following section, we give an overview of various works related to this research. Section 3 contains in-depth information about the $AutoFAQ$ system that we propose, including details about the summarization, question generation, and question answering systems that we implement and the heuristics that we employed for filtering purposes. Section 4 contains information about our evaluation method for evaluating $AutoFAQ$ system. In section 5, we present the results of the experiment. Finally, section 6 and 7 consist of conclusion and future work, respectively.

---

[1] SEP is available at https://plato.stanford.edu/
[2] Kaggle Dataset: https://www.kaggle.com/dataset/4042b656772a95f85e63ebc7438d8e91c86228aa76c9e64c536be7a39ee3fa54 or short link: http://bit.ly/platodata

## 2 RELATED WORK

In this section, we give an overview of the recent related works in the domains of text summarization, question answering, and question generation.

### 2.1 Summarization

Text summarization usually involves gathering all the important information from a text document, making it easier to understand the concepts in it. One can define summarization as the act of expressing the most important facts or ideas about a text in a short and clear form. The automatic text summarization techniques can be broadly classified into extractive and abstractive summarization, where the former is often defined as a binary classification task with labels indicating whether a text span should be included in the summary or not, while the latter involves introducing out-of-vocabulary words to improve the quality of summary making it relative to summarization done by humans. In this work, we implement the model of the pre-trained encoder developed in Liu et al. [9] available through the open-source transformer model provided by HuggingFace [21]. Liu et al. [9] explore the potential of BERT language model for text summarization under a general framework encompassing both extractive and abstractive modeling paradigms. Their extractive model is built on top of document level encoder based on BERT which encodes a given document and obtains representations for its sentences. Following this, an extractive model is built on top of the encoder by stacking several inter-sentence transformer layers to capture document-level features for extracting sentences.

Some previous works for summarization involved the use of neural networks such as using hierarchical document encoder with attention-based decoder (Cheng et al. 2016 [2]), and RNN based sequence model (Nallapati et al. 2017 [11]). Most recently, Zhong et al. (2020) [24] formulated the task of summarization as a semantic text matching problem, where the source document and candidate summaries are matched in a semantic space. We use the pretrained encoders model developed by Liu et al. (2019) [9] for two reasons. One reason is that using the similar embeddings from BERT in all the stages could be beneficial. The other reason is that we'll be able to exploit the advantages of employing both abstractive and extractive summarization in the process.

### 2.2 Question Generation

After the first question generation shared task evaluation challenge Rus et al. (2010) [16], that took place in Spring 2010, question generation task has received a wide spread attention in the natural language generation community. Some approaches involved creating robust frame works and templates, and using them to generate the questions. Heilman et al. (2010) [5] present an approach that addresses the challenge of automatically generating questions from reading materials for educational practice and assessment. Their method is quite different from our work, because they over-generate questions using hard coded rules and then use logistic regression to rank the questions, whereas we use cosine similarity and Text Rank paradigms for ranking the sentences before generating questions. We chose this approach because our hypothesis depends on the assumption that good FAQs should be covering important concepts of the document. Labutov al.(2015) [7] generated high-level comprehension questions from novel text that can bypass the myriad challenges of creating a full semantic representation. They exploit the low-dimensional ontology of document segments from original text and then use a set of question templates that are crowd sourced and matched with that representation while simultaneously ranking them based on relevance to the source. Around the same time, Chali et al. (2015) [1] considered the automatic generation of all possible questions from a topic of interest by exploiting the named entities and predicate argument structures of sentences. Many approaches in the recent years are based on the neural encoder-decoder architectures. Serban et al. (2016) [17] tackle the problem by transducing the knowledge graph facts into questions. They create a factoid question and answer corpus by using RNNs. There is also some work that combines question generation with question answering for further improvement. On the contrary to such works, our corpus being philosophical documents do not involve much of factual information but rather they have high degree of ideologies. Tang et al. (2017) [18] considered these tasks of question generation and question answering as dual tasks and trained their relative models simultaneously. Their training uses the probabilistic correlation between those two tasks. The question generation task is also used in computer vision where questions are generated from the given images Mostafazadeh et al. (2016) [10]. One of the latest studies conducted by Du et al. (2017) [4] involves question generation on both sentence and paragraph levels for the purpose of reading comprehension tasks, and they adopt an attention based sequence learning model. Another recent work is by Yuan et al. (2017) [22], they generate questions from documents using supervised and reinforcement learning. For question generation from sentences we implemented the model based on Kriangchaivech et al. [6] using the open source code provided by the author at GitHub[20].

### 2.3 Question Answering

Devlin et al.[3] designed BERT to pre-train deep bidirectional language representations by jointly training on both left and right contexts of a given word in all layers of their model. BERT's model architecture is a multi-layer bidirectional transformer encoder, originally implemented by Vaswani et al. (2017) [19]. Furthermore, the work of Devlin et al. [3] follows from a long history of pre-training general language representation models. Specifically, GloVe embeddings developed by Pennington et al. (2014) [14], which rely on word-to-word co-occurrence statistics, and ELMo embeddings developed by Peters et al. (2017) [15], which generalize traditional word embeddings by integrating context-sensitive features from language models. Nonetheless, pre-trained embeddings have become integral to NLP systems, offering significant improvements over embeddings learned from scratch. Moving forward, Devlin et al. [3] achieved state-of-the-art results on a wide range of NLP tasks by fine-tuning BERT. Fine-tuning approaches involve pre-training some model architecture on a language model objective before fine-tuning that same model for a supervised downstream task. With a fine-tuning approach, models only need to learn a few parameters from scratch. This is why we used a question answering system pre-trained with BERT embeddings on SQuAD 1.0 and SQuAD 2.0 datasets.

## 3 SYSTEM

The system diagram in Figure **??** represents 3 major modules that we use to generate the questions and 4 heuristics that we use to filter out the questions. First, we got the HTML data and processed it into sections and paragraphs for each article accordingly. Then we summarize the articles at paragraph-level. After we get the summarized paragraphs, we use the TextRank algorithm to rank all the sentences with their salience, and sort them in the descending order of their saliency. From this sorted list of sentences, we generate questions for top 100 sentences. We ignore the rest of the sentences because we assume that their salience would be relatively lower than that of the chosen sentences. We also consider having more than one possible question for each sentence, and we remove duplicate questions. On this set of questions and their respective answers, we perform filtering using some heuristics so that bad and invalid questions are systematically ignored, and more plausible questions are selected for evaluation. More information about heuristics used can be found in section 3.4.

### 3.1 Summarization

For summarization, we implement the BERT transformers from Devlin et al.[3] and the model from [9] which has both abstractive and extractive summarizations. The model is based on a document level encoder on BERT embeddings which is able to express the semantics of the documents and represent the sentences. The encoder is built on this by stacking inter sentence transform layers. For abstractive summarization, a new model is proposed with fine tuning and scheduling which adopts different optimizers for the encoder and decoder as a means of alleviating the mismatch between them. Fine-tuning with BERT form summarizations is not straight forward. BERT has the output as tokens but while extractive summarization is sentence level representations. The BERT-SUM model in liu et al. [9] is shown in Figure 1.

e

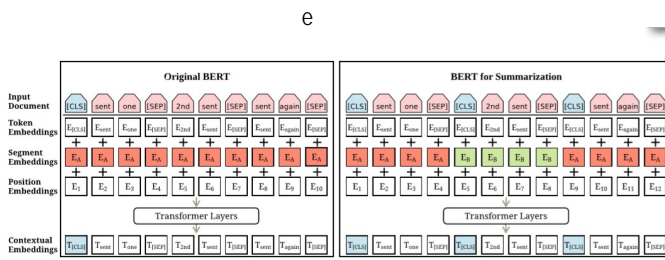

**Figure 1: BERTSUM from Liu et al.**

The individual CLS tokens are used to represent the sentence level structure. EA and EB are also used as tokens to represent the before and after sentences. For example, for document [sent1, sent2, sent3, sent4, sent5], we would assign embeddings [$EA$ , $EB$ , $EA$ , $EB$ , $EA$ ]. This way, document representations are learned hierarchically where lower transformer layers represent adjacent sentences, while higher layers, in combination with self-attention, represent multi-sentence discourse. Position embeddings in the original BERT model have a maximum length of 512; we overcome this limitation by adding more position embeddings that are initialized randomly and fine tuned with other parameters in the encoder.

This system is evaluated on CNN/NewYorktimes and Xsum data sets and Rouge-1, Rouge 2 and Rouge-L scores are calculated. It achieved state of the art or close results on all the three data sets.

### 3.2 Text Rank

The input to this module is a summarized article which is split in to sentences. The steps involved are

- Get the vector representations using GloVe embeddings.
- The cosine similarity score is calculated for all the sentences in the form of the matrix.
- A graph is formed with sentences as Vertices and Similarity Score as Edges.
- This is input to a Page Rank algorithm and the sentences are ranked to form the summary.
- The sentences are sorted in Ascending order according to the score and top 100 Sentences are Used.

*PageRank* [13] algorithm used in Text Rank is usually used for ranking the web pages based on the links it has with other pages. It calculates the score based on the probability that a user can reach a web page from the current web page and sums up the score for the current web page to all other web pages to get its total score which is used for the ranking. We used networkx module that provided a generic version of page rank that takes undirected graphs and can be customized to any scenario. The eigenvectors[8] might not converge in this case over the iterations but we stopped the Page rank at the max iterations. In our case the score is the similarity score and the pages are the sentences.

### 3.3 Question Generation

From the ranked sentences sorted in the descending order of their importance, we generate questions for each of the sentences. For reduced use of computational resources, we only consider the top 100 highly ranked sentences at this step. We employ the Question Generation by Transformers model developed by Kriangchaivech et al. (2019) [6] for this step in our system. We use the code available at [20]. This involves a seq2seq model to automatically generate questions from a given passage. The input passage can be as small as a single sentence and we consider more than one possible question for a given sentence. We chose to employ this method because it is based on transformers rather than RNN, so parallelized inference can be achieved. Transformers also incorporate the beam search and bucketing mechanisms. The transformers used have an encoder and a decoder and also adopt multiple attention heads which is not quite possible in RNNs. The model that we used in our work [6] is trained on the Stanford Question Answering Dataset (SQuAD) which consists of 100,000+ questions posed on Wikipedia articles, where the answer to each question is a segment of text from the corresponding reading passage. Their model's evaluation is done by calculating the word error rate (WER) between the generated questions and the corresponding questions from SQuAD dataset. In our work, we generate 2 questions for each of the top 100 highly ranked sentences, and ignore the duplicate questions. We do this because, there can be more than one valid question posed on a certain sentence, and it can be useful to determine them. In further steps, we filter out the questions as explained in the following section.

## 3.4 Heuristics

This section explains briefly about the heuristics that we employ while filtering the questions.

*3.4.1 Heuristic 0 - Comparison with QA system.* [3] We used a QA system trained on SQuAD created by Devlin et. al. [3] provided in the HuggingFace transformers python package[21] that achieved a state-of-the-art F1 score of around 92 on SQuAD 1.0 and around 84 on SQuAD 2.0.

Since we generate questions based on a span (the answer) within an input sentence, we suppose that a good QA system would find the same answer when given the generated question and the original sentence (as the context for the question). If we find a match between our expected answer and the QA system's output answer then we keep our generated question.

*3.4.2 Heuristic 1 - Grammar Check.* We pass our question into a grammar checker using the GrammarBot API. We allow TYPOGRAPHY mistakes (e.g. white-space proceeding a comma) because such mistakes are trivial to fix. We disallow all other grammar mistakes. The GrammarBot API that we have employed for this heuristic can be found at: https://pypi.org/project/grammarbot/

*3.4.3 Heuristic 2 - Multiple WH-words.* We identify WH-words from the generated question using the en_core_web_sm 2.2.5 model included in the SpaCy python package. We look for "WDT","WP", "WP$", "WRB", and remove those questions that have more than one such identifier in them. The details of the model we employed in this step can be found at the url: https://spacy.io/models/en

*3.4.4 Heuristic 3 - Named Entity Recognition.* We extract named entities from the generated question and ensure that the questions that contain named entities are given more preference, as they tend to have more factual content in them. Again, we use SpaCy's named entity recoginition model to extract entities. More information on how the above mentioned heuristic rules have been employed in our work can be found in the Kaggle notebook. [4]

## 4 EVALUATION

We performed human evaluations to evaluate the proposed our frequently asked questions generation system. After the questions are filtered using the heuristics mentioned in the previous section, we choose the top 25 question of each document for evaluation. Each question in this set is evaluated by 3 individuals based on the following criteria:

- How was the grammar of the question? – Rate the quality of grammar of the given question. The rating can be from 0 to 5 stars where 0 corresponds to bad and 5 corresponds to excellent.

- Is the question meaningful? – Determine if the given question is related to document and makes sense. The rating can be Yes, No, or Maybe. If in case Maybe is selected a reason must be given.
- How well did the paragraph answer the question? – Rate the accuracy of the provided answer with the corresponding question. The ratings are from 0-5 stars where 0 corresponds to bad and 5 corresponds to excellent.

## 4.1 Evaluation strategy

Each participant was presented with a web-page that greeted then with a pseudonym stored within their browser window.

Then the participants would see a link to the original article in the Stanford Encyclopedia of Philosophy website.

Then the participants would see a single generated question.

Then the participants would see a summarized paragraph.

We made sure that every question would be evaluated by 3 participants using database constraints and views within AirTable for the purpose of taking the average of ratings as the evaluation metric of our system. [5]

## 5 RESULTS

Overall, we rated 100 questions. The three of us submitted the most evaluations. Some other individuals also provided their perspectives on a few questions. Some examples of the questions rated are given below. The articles titled "Cancer" "Copenhagen Interpretation of Quantum Mechanics" "Cusanus, Nicolaus [Nicolas of Cusa]" and "Evolutionary Epistemology" were completely evaluated (three human ratings per question). We also generated 12 FAQ documents each with 25 questions. [6] Table 4 shows the results that the system has achieved on human evaluation. As the grammar and how good the presented answer answers the question are rated between 1-5, we find that the grammar of most of the questions on all 4 evaluated documents was pretty good. We also find a correlation between the grammar and answerability where, when the participants think that the answer presented is correct, then they were convinced that the grammar of the question is also plausible in it's own right. This correlation can be observed from Figure 3. During the evaluation, we find certain questions to be too generic like the one displayed in Table 1. If a user is presented with such a question then it would only confuse him more rather than answering his questions. At the same time, we also find some really good questions generated as shown in Tables 2 and 3. These questions are related to the context available in their respective paragraphs. One major weakness we find in many questions is that there was no proper co-reference resolution done, so some of the questions could've been really good if they had the right referencing involved.

---

[3] Source https://github.com/huggingface/transformers
[4] https://www.kaggle.com/teja0110/filtered-questions
[5] Human Evaluation Website: https://582eval.glitch.me/
[6] More AutoFAQ documents found at github.com/nagasaichandra/AutoFAQ
[7] plato.stanford.edu/entries/cancer/

[8] plato.stanford.edu/entries/cusanus/

[9] plato.stanford.edu/entries/cusanus/

| Cancer |
|---|
| **What has many causes , and many effects?** |
| *summarized_paragraph* : The history of attempts to identify defining features of cancer, let alone arrive at a unified theory, has floundered. Either cancer is defined so vaguely as to include non-pathological states, or focused so narrowly on a specific class of causes. **Such a variety of definitions is due to the fact that cancer has many causes, and many effects.** It involves many different types of dysregulation, at a variety. of temporal and spatial scales. It is no small challenge to identify defines cancer, and arrive at unified "theory" of disease etiology. |
| *avg_grammar_rating* : 4.3 |
| *avg_answerability_rating* : 5.0 |
| *sum_yes_meaningful* : 1 |
| *sum_no_meaning* : 1 |
| *sum_maybe_meaning* : 1 |

**Table 1: An example FAQ from "Cancer" article**

| Cusanus, Nicolaus [Nicolas of Cusa] |
|---|
| **How is the presence of god and god's identity with things not to be thought of as?** |
| *summarized_paragraph* : Christian Neoplatonism is distinctive for its ability to hold together dialectically in thought the insight it provides about this asymmetrical, non-reciprocal connection between God and creatures. God penetrates and surpasses or exceeds each thing God creates and encompasses. Creatures are thus themselves real with the limited sort of independence they manifest, yet they are at once in God and indeed one with God without being themselves divine. **This means that the presence of God and God's identity with things is not to be thought as the kind of reciprocity, say, that two created physical things have.** |
| *avg_grammar_rating* : 5.0 |
| *avg_answerability_rating* : 5.0 |
| *sum_yes_meaningful* : 3 |
| *sum_no_meaning* : 0 |
| *sum_maybe_meaning* : 0 |

**Table 2: An example FAQ from "Cusanus, Nicolaus" article**

## 6 CONCLUSION

Overall, the methodology we used is able to generate some frequently asked questions and answers and the evaluation showed that the questions generated have an average grammar rating of 4.38/5 and the tagged answers had an average rating of 3.81/5. Overall 71.28% of questions were rated to be meaningful. These results show that the proposed model is able to generate some very good question which represent the context of the paragraph. At the same time the model has some vague questions which are difficult to comprehend. The system could generate few nonsensical questions but we have managed to filter most of them out using the various heuristics.

| Cusanus, Nicolaus [Nicolas of Cusa] |
|---|
| **What is the wall of paradise?** |
| *summarized_paragraph* : Cusanus leads us through a series of reflections on seeing and on the face of God. God is located beyond both imaginative exercise and conceptual understanding. **Nicholas symbolizes our approach to this beyond by encouraging us to enter "into a certain secret and hidden silence wherein there is no knowledge or concept of a face," characterizing it as an "obscuring mist, haze, darkness or ignorance." He invokes the coincidence of opposites from On Learned Ignorance and proposes his second central metaphor: the wall of paradise.** |
| *avg_grammar_rating* : 5.0 |
| *avg_answerability_rating* : 4.3 |
| *sum_yes_meaningful* : 3 |
| *sum_no_meaning* : 0 |
| *sum_maybe_meaning* : 0 |

**Table 3: An example FAQ from "Cusanus, Nicolaus" article**

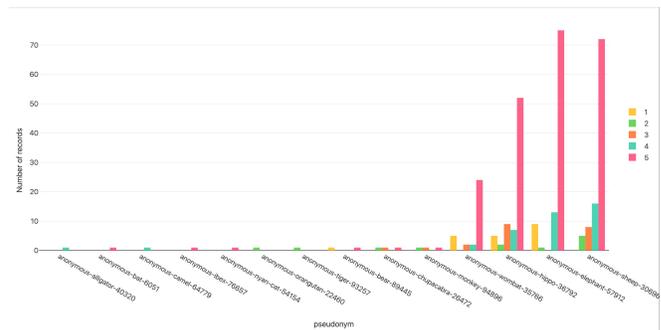

**Figure 2: Pseudonyms grouped by grammar rating.**

| Article | No. | Avg Grammar | Avg Answerability | Meaningfulness |
|---|---|---|---|---|
| 1 | 79 | 4.06 | 3.41 | 63.29 |
| 2 | 85 | 4.54 | 3.89 | 72.94 |
| 3 | 81 | 4.33 | 4.12 | 76.54 |
| 4 | 76 | 4.59 | 3.84 | 72.36 |
| **Total** | **321** | **4.38** | **3.81** | **71.28%** |

**Table 4: Average ratings of grammar and answerability, percentage of questions rated to be meaningful**

## 7 FUTURE WORK

The proposed model Produced some good questions which a good grammar sense. At the same time there could be some improvements done to remove vague questions. This includes adding co-reference resolution which could solve some questions which have good Answers but the question is either vague or . The Future work Could also include a Corpora of Tagged entities that are related to the Corpora that the Questions are generated for. This could resolve

| Article | Title |
|---|---|
| 1 | Cancer |
| 2 | Copenhagen Interpretation of Quantum Mechanics |
| 3 | Cusanus, Nicolaus [Nicolas of Cusa] |
| 4 | Evolutionary Epistemology |

Table 5: Articles

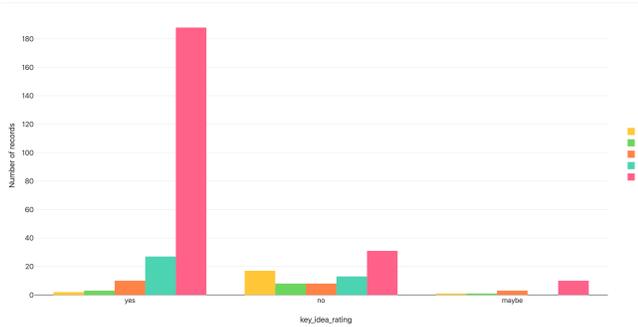

Figure 3: Meaningfulness (key_idea) grouped by grammar rating.

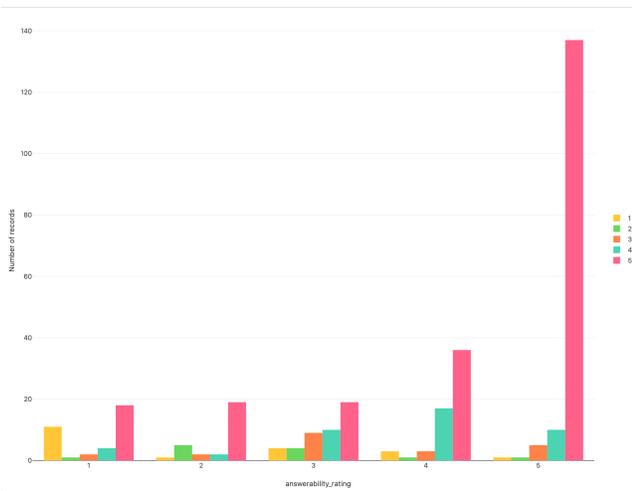

Figure 4: Answerability grouped by grammar rating.

some issues like misinterpretation of entities that have multiple words.

## ACKNOWLEDGMENTS

We would like to acknowledge Dr. Foaad Khosmood for sharing his great depth of knowledge in natural language processing and for helping us scrape the philosophical articles.

## REFERENCES
[1] Yllias Chali and Sadid A Hasan. 2015. Towards topic-to-question generation. *Computational Linguistics* 41, 1 (2015), 1–20.
[2] Jianpeng Cheng and Mirella Lapata. 2016. Neural summarization by extracting sentences and words. *arXiv preprint arXiv:1603.07252* (2016).
[3] Jacob Devlin, Ming-Wei Chang, Kenton Lee, and Kristina Toutanova. 2018. Bert: Pre-training of deep bidirectional transformers for language understanding. *arXiv preprint arXiv:1810.04805* (2018).
[4] Xinya Du, Junru Shao, and Claire Cardie. 2017. Learning to ask: Neural question generation for reading comprehension. *arXiv preprint arXiv:1705.00106* (2017).
[5] Michael Heilman and Noah A Smith. 2010. Good question! statistical ranking for question generation. In *Human Language Technologies: The 2010 Annual Conference of the North American Chapter of the Association for Computational Linguistics*. Association for Computational Linguistics, 609–617.
[6] Kettip Kriangchaivech and Artit Wangperawong. 2019. Question Generation by Transformers. *arXiv preprint arXiv:1909.05017* (2019).
[7] Igor Labutov, Sumit Basu, and Lucy Vanderwende. 2015. Deep questions without deep understanding. In *Proceedings of the 53rd Annual Meeting of the Association for Computational Linguistics and the 7th International Joint Conference on Natural Language Processing (Volume 1: Long Papers)*. 889–898.
[8] Amy N Langville and Carl D Meyer. 2005. A survey of eigenvector methods for web information retrieval. *SIAM review* 47, 1 (2005), 135–161.
[9] Yang Liu and Mirella Lapata. 2019. Text summarization with pretrained encoders. *arXiv preprint arXiv:1908.08345* (2019).
[10] Nasrin Mostafazadeh, Ishan Misra, Jacob Devlin, Margaret Mitchell, Xiaodong He, and Lucy Vanderwende. 2016. Generating natural questions about an image. *arXiv preprint arXiv:1603.06059* (2016).
[11] Ramesh Nallapati, Feifei Zhai, and Bowen Zhou. 2017. Summarunner: A recurrent neural network based sequence model for extractive summarization of documents. In *Thirty-First AAAI Conference on Artificial Intelligence*.
[12] Preksha Nema and Mitesh M Khapra. 2018. Towards a better metric for evaluating question generation systems. *arXiv preprint arXiv:1808.10192* (2018).
[13] Lawrence Page, Sergey Brin, Rajeev Motwani, and Terry Winograd. 1999. *The pagerank citation ranking: Bringing order to the web.* Technical Report. Stanford InfoLab.
[14] Jeffrey Pennington, Richard Socher, and Christopher D Manning. 2014. Glove: Global vectors for word representation. In *Proceedings of the 2014 conference on empirical methods in natural language processing (EMNLP)*. 1532–1543.
[15] Matthew E Peters, Waleed Ammar, Chandra Bhagavatula, and Russell Power. 2017. Semi-supervised sequence tagging with bidirectional language models. *arXiv preprint arXiv:1705.00108* (2017).
[16] Vasile Rus, Brendan Wyse, Paul Piwek, Mihai Lintean, Svetlana Stoyanchev, and Cristian Moldovan. 2010. The first question generation shared task evaluation challenge. (2010).
[17] Iulian Vlad Serban, Alberto García-Durán, Caglar Gulcehre, Sungjin Ahn, Sarath Chandar, Aaron Courville, and Yoshua Bengio. 2016. Generating factoid questions with recurrent neural networks: The 30m factoid question-answer corpus. *arXiv preprint arXiv:1603.06807* (2016).
[18] Duyu Tang, Nan Duan, Tao Qin, Zhao Yan, and Ming Zhou. 2017. Question answering and question generation as dual tasks. *arXiv preprint arXiv:1706.02027* (2017).
[19] Ashish Vaswani, Noam Shazeer, Niki Parmar, Jakob Uszkoreit, Llion Jones, Aidan N Gomez, Łukasz Kaiser, and Illia Polosukhin. 2017. Attention is all you need. In *Advances in neural information processing systems*. 5998–6008.
[20] Artit Wangperawong. 2020. Text2Text: generate questions and summaries for your texts. https://github.com/artitw/text2text. https://github.com/artitw/text2text
[21] Thomas Wolf, Lysandre Debut, Victor Sanh, Julien Chaumond, Clement Delangue, Anthony Moi, Pierric Cistac, Tim Rault, R'emi Louf, Morgan Funtowicz, and Jamie Brew. 2019. HuggingFace's Transformers: State-of-the-art Natural Language Processing. *ArXiv* abs/1910.03771 (2019).
[22] Xingdi Yuan, Tong Wang, Caglar Gulcehre, Alessandro Sordoni, Philip Bachman, Sandeep Subramanian, Saizheng Zhang, and Adam Trischler. 2017. Machine comprehension by text-to-text neural question generation. *arXiv preprint arXiv:1705.02012* (2017).
[23] Edward N Zalta, Uri Nodelman, Colin Allen, and R Lanier Anderson. 1995. Stanford encyclopedia of philosophy. https://stanford.library.sydney.edu.au/index.html. https://stanford.library.sydney.edu.au/index.html
[24] Ming Zhong, Pengfei Liu, Yiran Chen, Danqing Wang, Xipeng Qiu, and Xuanjing Huang. 2020. Extractive Summarization as Text Matching. *arXiv preprint arXiv:2004.08795* (2020).